# FRACCO: A gold-standard annotated corpus of oncological entities with ICD-O-3.1 normalisation

## Abstract

Developing natural language processing tools for clinical text requires annotated datasets, yet French oncology resources remain scarce. We present FRACCO (FRench Annotated Corpus for Clinical Oncology) an expert-annotated corpus of 1'301 synthetic French clinical cases, initially translated from the Spanish CANTEMIST corpus as part of the FRASIMED initiative.

Each document is annotated with terms related to morphology, topography, and histologic differentiation, using the International Classification of Diseases for Oncology (ICD-O) as reference. An additional annotation layer captures composite expression-level normalisations that combine multiple ICD-O elements into unified clinical concepts.

Annotation quality was ensured through expert review: 1'301 texts were manually annotated for entity spans by two domain experts. A total of 71'127 ICD-O normalisations were produced through a combination of automated matching and manual validation by a team of five annotators. The final dataset representing 399 unique morphology codes (from 2'549 different expressions), 272 topography codes (from 3'143 different expressions), and 2'043 unique composite expressions (from 11'144 different expressions).

This dataset provides a reference standard for named entity recognition and concept normalisation in French oncology texts.

## Authors & Affiliations

Johann PIGNAT*[1,2], Milena VUCETIC*[1], Christophe GAUDET-BLAVIGNAC[1], Jamil ZAGHIR[1], Amandine STETTLER[1], Fanny AMREIN[2], Jonatan BONJOUR[2], Jean-Philippe GOLDMAN[1], Olivier MICHIELIN[2], Christian LOVIS[1], Mina BJELOGRLIC[1]

* Both the first and second authors can be considered co-first authors.

Affiliations:

1) Service des sciences de l'information médicale, Hôpitaux Universitaires de Genève, Suisse
2) Service d'oncologie de précision, Hôpitaux Universitaires de Genève, Suisse

## Background & Summary

Most clinical information contained within electronic health records (EHR) is in unstructured, free-text format (1). Extracting structured representations from such narrative textual data is important for both medical research and clinical practice. Natural language processing (NLP) tools, including named entity recognition (NER), are used for that purpose (2). However, their development, evaluation and validation depend crucially on the availability of annotated datasets (3). While significant efforts have been directed towards generating annotated datasets from English-language medical reports (4–8) , there remains a lack of high-quality, manually annotated French-language corpora (9).

To address this, several initiatives have emerged to create French-language clinical corpora with semantic annotations. The MERLOT corpus (10), for instance, offers a richly annotated dataset encompassing various medical entities and relations, facilitating research in clinical NLP for French texts. Similarly, the QUAERO French Medical Corpus (11) provides annotations for NER and normalisation tasks, contributing to the development of NLP tools tailored to French biomedical texts. The SIFR Annotator project (12) has also advanced the field by enabling ontology-based semantic annotation of French biomedical text, leveraging resources like the French ICD-10.

Despite these efforts, resources specifically focused on oncology and utilising oncology-specific ontologies for annotation remain scarce in the French language. Our dataset addresses this gap by providing a comprehensive, French-language corpus of synthetic clinical cases in oncology, annotated with ICD-O codes for morphology, topography, and differentiation. This resource not only complements existing corpora but also introduces expression-level normalisations, capturing complex clinical expressions that combine multiple ICD-O components. By presenting this dataset, we aim to support the development and evaluation of NLP applications in oncology, particularly for French-language clinical data, and to facilitate cross-lingual research by aligning and improving upon resources like the Spanish-language CANTEMIST (13) corpus.

The dataset includes 71'065 total entity annotations, divided into four categories: *morphologie*, *topographie*, *differenciation*, and a high-level composite label, *expression_CIM*, which groups multiple ICD-O components into a single unified clinical expression. The distribution of annotations across these categories and their relative proportions are summarised in *Figure 1*. The ditributions of the the ten most common full expression codes are shown in *Figure 2*.

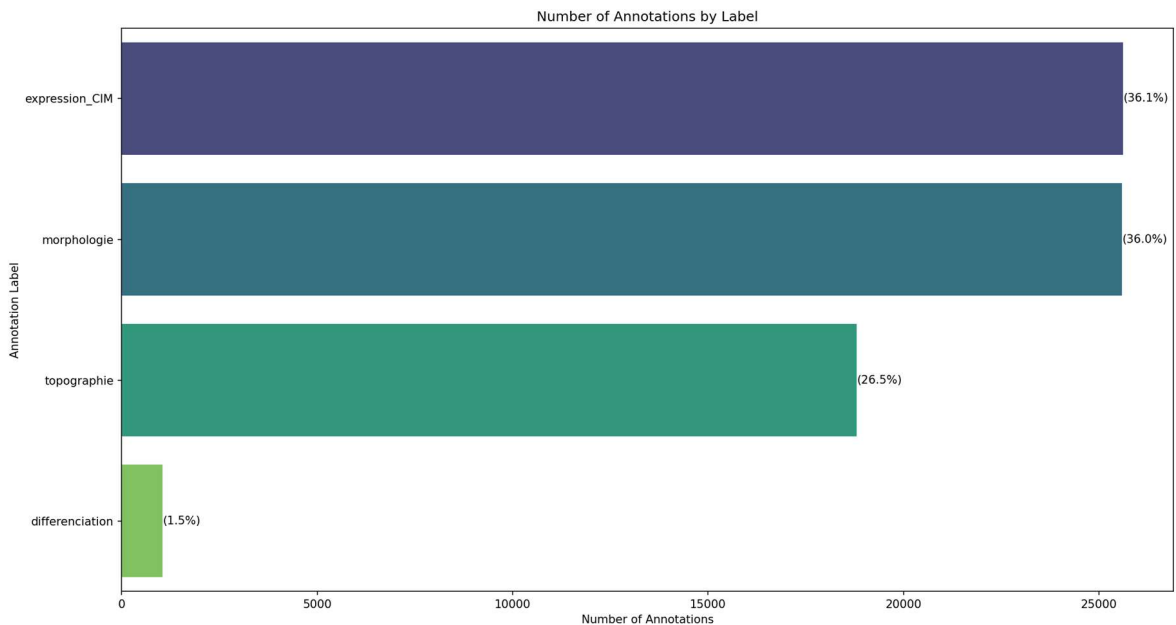

*Figure 1: label distribution, n = 71'065*

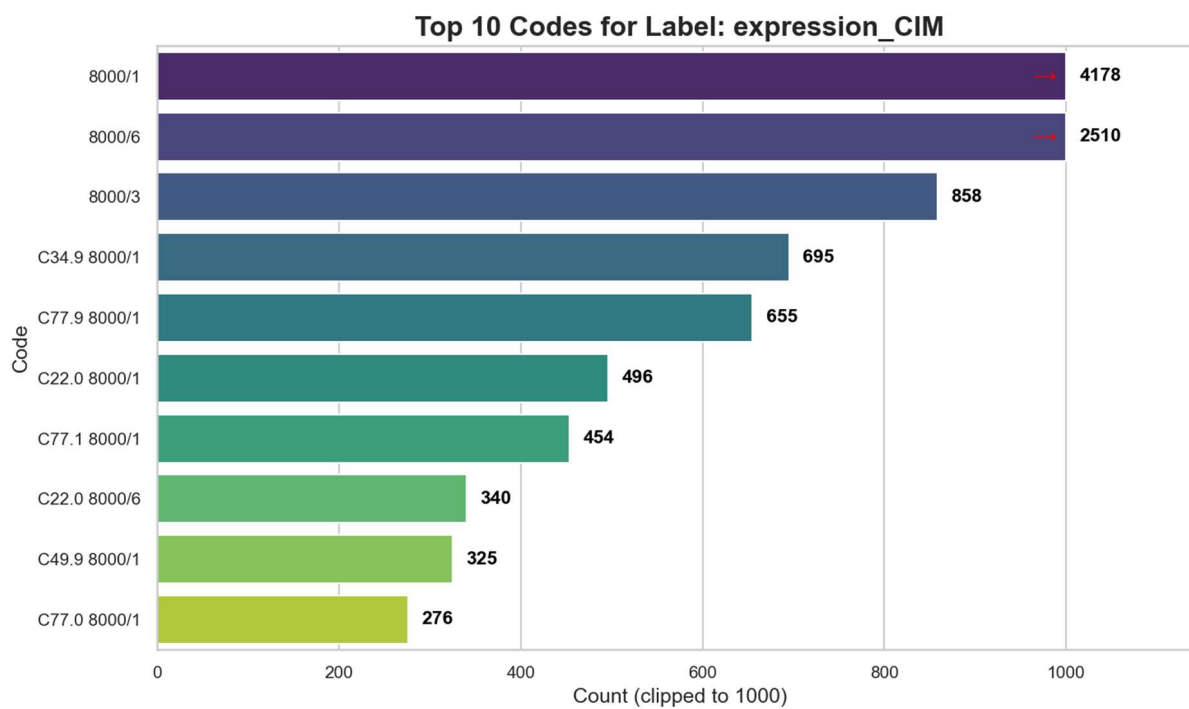

*Figure 2: most common expression_CIM codes: **8000/1** (Neoplasm, uncertain whether benign or malignant), **8000/6** (Neoplasm, metastatic), **8000/3** (Neoplasm, malignant), **C34.9** (Lung, NOS), **C77.9** (Lymph node, NOS), **C22.0** (Liver), **C77.1** (Intrathoracic lymph nodes), **C49.9** (Connective, subcutaneous and other soft tissues, NOS), **C77.0** (Lymph nodes of head, face and neck). NOS: « not otherwise specified »*

Each annotated entity was also normalised using the ICD-O-3 terminology. For atomic entities, exact ICD-O codes are provided. For expression-level annotations, composite codes combining *morphologie* (*Figure 3*), *topographie* (*Figure 4*), and *differenciation* (*Figure 5*) were assigned. The ten most frequently occurring codes in each annotation category are presented in *Figure 3-5* respectively, highlighting the clinical concepts most represented in the corpus and offering insight into its coverage.

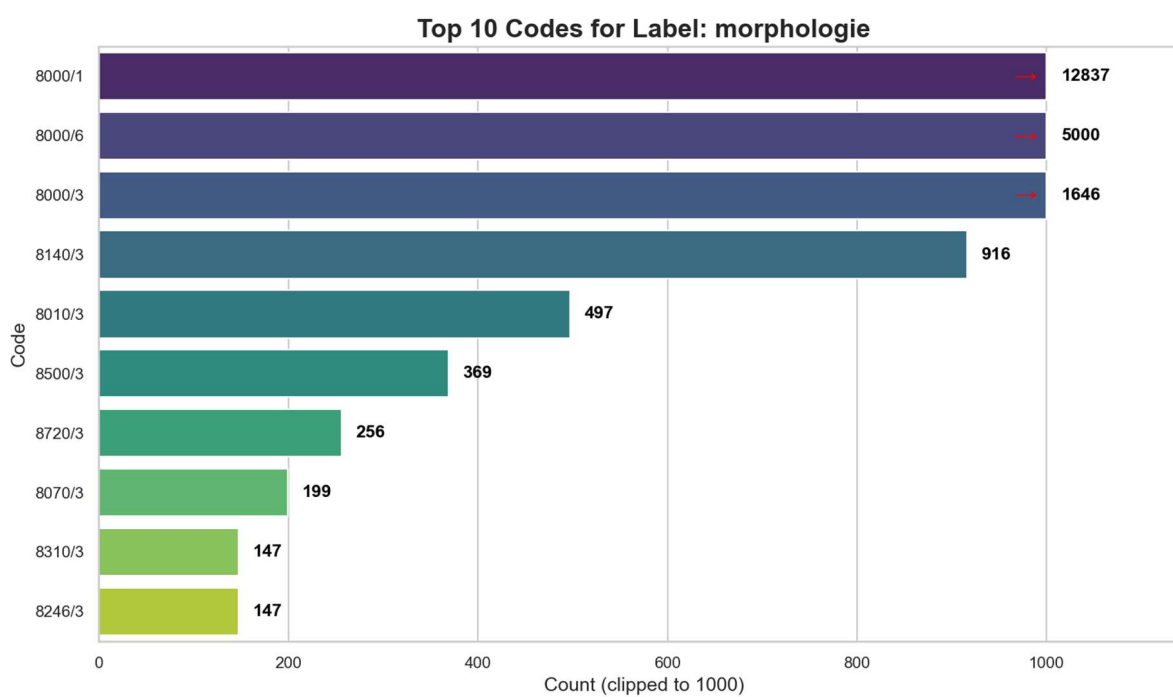

*Figure 3: most common morphologie codes: **8000/1** (Neoplasm, uncertain whether benign or malignant), **8000/6** (Neoplasm, metastatic), **8000/3** (Neoplasm, malignant), **8140/3** (Adenocarcinoma, NOS), **8010/3** (Carcinoma, NOS), **8500/3** (Infiltrating*

*duct carcinoma, NOS)*, **8720/3** *(Malignant melanoma, NOS)*, **8070/3** *(Squamous cell carcinoma, NOS)*, **8310/3** *(Clear cell adenocarcinoma, NOS)*, **8246/3** *(Neuroendocrine carcinoma, NOS). NOS: « not otherwise specified »*

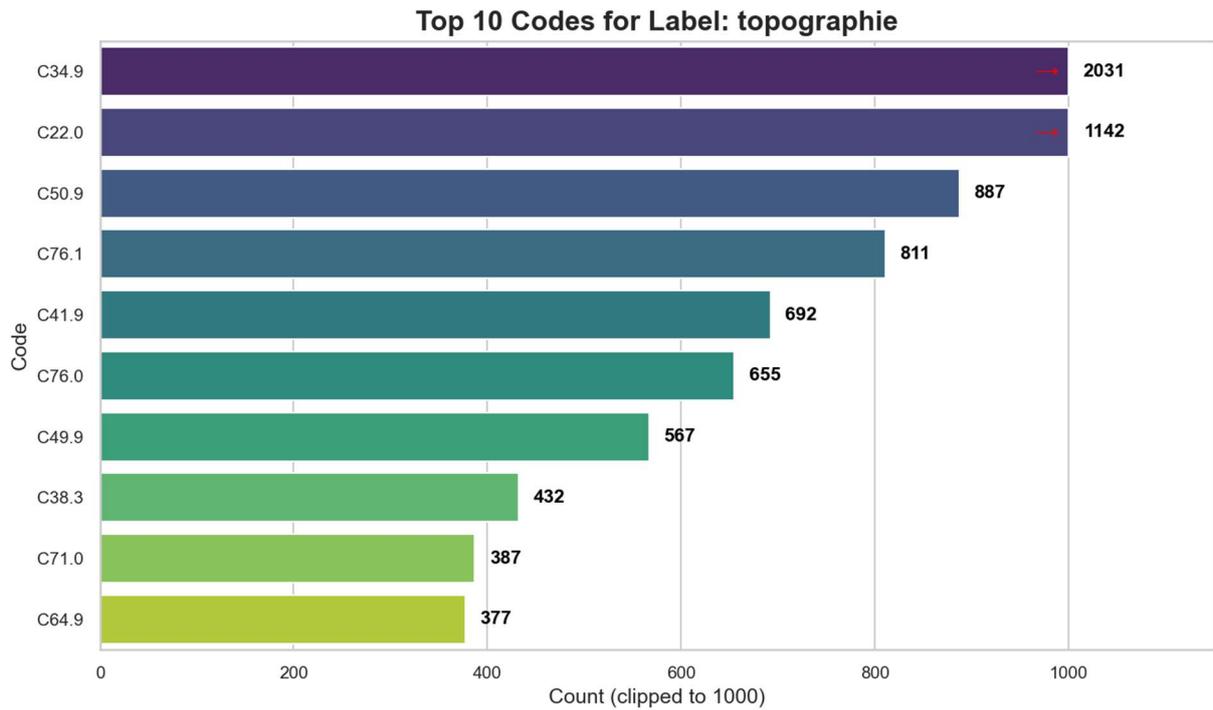

*Figure 4: most common topographie codes:* **C34.9** *(Lung, NOS),* **C22.0** *(Liver),* **C50.9** *(Breast, NOS),* **C76.1** *(Thorax, NOS),* **C41.9** *(Bone, NOS),* **C76.0** *(Head, face or neck, NOS),* **C49.9** *(Connective, subcutaneous and other soft tissues, NOS),* **C38.3** *(Mediastinum, NOS),* **C71.0** *(Cerebrum),* **C64.9** *(Kidney, NOS). NOS: « not otherwise specified »*

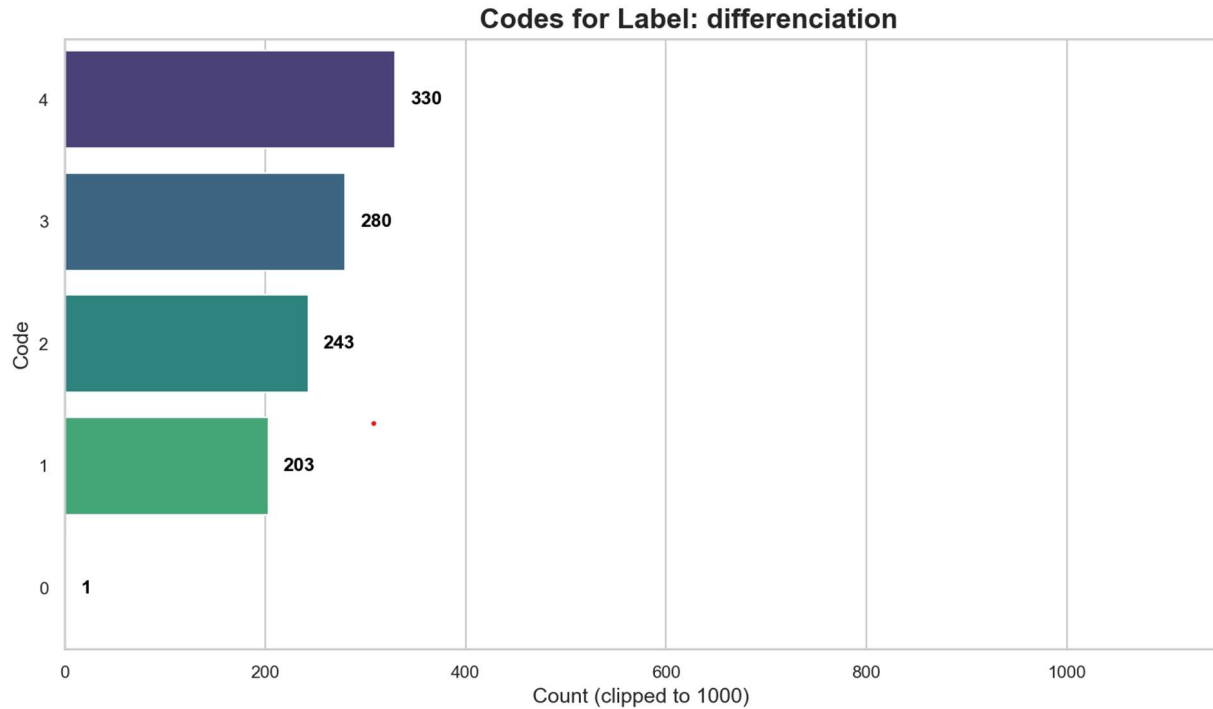

*Figure 5: disribution of differenciation codes:* **4** *(Undifferentiated, grade IV),* **3** *(Poorly differentiated, grade III),* **2** *(Moderately differentiated, grade II),* **1** *(Well differentiated, grade I),* **0*** *("grade 0"). *Non- standard.*

# Methods

## Source text corpus and translation

The FRACCO corpus presented in this dataset originates from the CANTEMIST corpus, a publicly available resource developed as part of the IberLEF 2020 challenge for named entity recognition and normalisation in oncology-related Spanish clinical texts. CANTEMIST consists of de-identified, synthetic clinical case reports focused on cancer diagnoses and terminology, annotated with morphology codes from the third edition of the International Classification of Diseases for Oncology (ICD-O-3).

To facilitate multilingual research and extend the usability of the CANTEMIST corpus to French-speaking contexts, the texts were translated from Spanish to French as part of the FRASIMED project (14). FRASIMED is a parallel corpus of synthetic clinical documents designed to support the development of clinical NLP tools across languages. The translation process was conducted using document-level machine translation (DeepL Pro).

The translated French texts served as the foundation for the present dataset. The CANTEMIST subset of 1'300 documents from FRASIMED was selected to reflect a broad range of oncological expressions and ICD-O morphological entities present in the original corpus.

## Annotation projection

The original Spanish CANTEMIST corpus contained manual annotations of oncology-related entities, specifically morphological terms, encoded with ICD-O-3 morphology codes. These annotations were produced using the BRAT annotation tool (15), which employs a standoff format where entity mentions and associated metadata (such as normalisation codes) are stored in .ann files alongside the raw text in .txt format.

To preserve the annotated structure, the original CANTEMIST annotations were automatically projected onto the French translations. This projection was performed using a direct alignment strategy based on sentence segmentation and character-level string matching. Since the texts were synthetic and structurally aligned, this method allowed for a relatively accurate transfer of annotations from the source to the target language.

However, due to lexical and syntactic differences between Spanish and French, a portion of the annotations required manual verification. In cases where phrase boundaries shifted during translation or where medical terms were rephrased, the projected annotations were reviewed and adjusted to ensure correct span alignment and semantic equivalence. The projected annotations thus served as a baseline for the extended annotation work conducted in the subsequent phases of the dataset construction.

## Refinement and extension

Following the initial projection of annotations onto the French-translated corpus, substantial effort was invested in this project to refine and extend the annotation layer to improve coverage, consistency, and semantic depth. While the original CANTEMIST annotations provided a strong foundation, they were limited in scope, covering only morphological oncology terms, and contained several inconsistencies, including missing entity mentions and incomplete normalisations.

Manual reannotation of the projected dataset addressed these shortcomings with a gold-standard quality. Annotators reviewed all documents to correct span alignment errors introduced during translation, and to recover relevant oncology terms that had been omitted in the original Spanish annotations. This process also involved harmonising entity boundaries and ensuring adherence to consistent annotation guidelines.

In addition to correcting and expanding on existing annotations, two new categories of ICD-O entities were introduced: topographical codes (describing the anatomical site of the tumour) and histology differentiation codes (indicating tumour grading). These were annotated de novo across the entire corpus using ICD-O standards, substantially enriching the dataset's expressiveness. Crucially, a new complete expression-level layer, termed "expression_CIM", was added to normalise complex expressions involving combinations of morphology, topography, and differentiation into unified entities. An example of the these can be found in *Figure 6*.

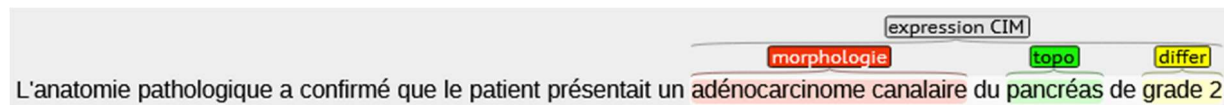

*Figure 6 : example of annotations labels*

These annotations capture clinically meaningful multi-attribute expressions (e.g., "carcinome épidermoïde bien différencié du poumon") and are each linked to their corresponding ICD-O codes. The creation of this layer enables higher-level semantic analysis and facilitates use cases such as relation extraction and document classification.

As a result of this extensive refinement and extension process, the total number of annotations in the corpus grew from approximately 16,000 in the original CANTEMIST dataset to over **70,000** in the current version. These annotations represent 2'549 unique *morphologie* expressions, 3'143 unique *topographie* expressions, 184 unique *différenciation* unique expressions, and 11'144 unique *expression_CIM* expressions. These unique expressions were mapped to a total of 399 ICD-O morphology codes, 375 topography codes, all four differentiation codes, and over 2'043 unique expression_CIM after normalisation. This expanded and corrected annotation layer provides a rich resource for French-language biomedical NLP and cancer information extraction tasks.

## ICD-O normalisation

To ensure semantic consistency and interoperability with standardised cancer ontologies, all entity annotations were normalised using the ICD-O-3. The normalisation process combined automated matching with manual review and was applied to both individual entity annotations ("morphologie", "topographie", "différenciation") and higher-level composite expressions (expression_CIM).

For single label terms and simple expressions, a dictionary-based matching approach was used. Each annotated span was compared via regular expressions to entries in a comprehensive ICD-O terminology lexicon. Exact matches were assigned the corresponding ICD-O code. When a term or expression already had a code from the original CANTEMIST annotations, the two codes were compared. Discrepancies between the dictionary-based match and the original code were flagged for manual review to verify correctness and address inconsistencies inherited from the source corpus. Terms that were not matched automatically were flagged for manual normalisation.

The construction of expression_CIM annotations involved combining individual entities into composite expressions representing full oncological concepts, such as "carcinome pulmonaire indifférencié" (*Table 1*). Expressions meeting simple criteria (containing one morphology, and optionally one topography and/or one differentiation), were automatically reconstructed from their components and assigned a composite annotation. More complex expressions, or those involving multiple entities per label, context dependent abbreviations or ambiguous phrasing, were flagged for manual and normalisation.

| Annotated text | Entity label | ICD-O3 code |
|---|---|---|
| **carcinome pulmonaire indifférencié** | expression_CIM | C34.9 8010/34 |
| **carcinome** | morphologie | 8010/3 |
| **pulmonaire** | topographie | C34.9 |
| **indifférencié** | différenciation | 4 |

*Table 1: composition of expression_CIM codes*

Special attention was given to expressions containing **"adénopathi-"** (e.g., *adénopathie*, *conglomérat adénopathique*, etc.). These cases often required a specific topography code (C77.-, referring to lymph nodes), but were context-dependent and not reliably annotated in the source data. To address this, expressions containing "adénopathi-" were programmatically pre-processed by substituting the morphological component with a generic equivalent, "ganglion." The resulting expression was then matched against the ICD-O topographical dictionary. If a valid match was found, the associated C77.- topography code was applied. Expressions for which no valid match could be constructed were set aside for manual review.

This semi-automated workflow allowed for high-coverage, consistent ICD-O normalisation at both the entity and expression level, balancing scalability with annotation quality. The result is a dataset enriched with structured oncology information suitable for downstream tasks in clinical NLP, semantic parsing, and ontology alignment.

## Correction of mistranslations

Because the synthetic clinical cases were translated from Spanish into French using the DeepL translation engine, some expressions were incorrectly rendered. This problem was particularly frequent with abbreviations and overly complex medical expressions, which were often left untranslated or were completely mistranslated, which was not representative of authentic French clinical language.

To address this, we implemented a systematic correction pipeline during the annotation process. First, annotators flagged mistranslated expressions when encountered in the corpus. These flagged strings were then used as seeds to automatically retrieve similar instances across the dataset. Each retrieved candidate was manually checked and flagged if necessary. In a second step, a Python script was developed to (i) retranslate flagged terms into appropriate French expressions, (ii) replace the incorrect spans directly in the text, and (iii) shift all existing annotations in the corresponding document to preserve alignment between the corrected text and the standoff annotation files.

This iterative process ensured that mistranslations were consistently corrected across the dataset. In total, 961 expressions were flagged and corrected, improving the linguistic quality and representativeness of the French corpus while maintaining annotation integrity.

## Annotation tools and workflow

Manual annotation was conducted using the Brat Rapid Annotation Tool (brat) (15), a web-based interface supporting span-based entity annotation in the standoff .ann format. Annotators used brat to revise projected annotations from the original CANTEMIST corpus, add new entities (morphology, topography, differentiation, full expressions), and ensure precise alignment with the corresponding French texts.

To support the normalisation process, all annotated entities were automatically extracted from the BRAT .ann files into a structured CSV file, where each entry could be independently reviewed and matched to the appropriate ICD-O code. This working format enabled efficient dictionary matching and allowed annotators to easily resolve inconsistencies and handle complex cases such as ambiguous or multi-entity expressions.

Annotators had access to the full original French text throughout the normalisation process. This allowed for context-sensitive interpretation of expressions. While many annotations could be assigned based on local lexical cues, certain expressions required broader contextual interpretation. For example, morphology expressions often carried implicit malignancy status (e.g., *adénome* vs. *adénocarcinome*), which was not always lexically specified. In some cases, context helped clarify whether a term referred to a *benign* (/0), *malignant* (/3), or *uncertain malignancy* (/1) process, particularly for more general or ambiguous terms such as *lésion*, *prolifération*, or even *tumeur*. Similarly, the interpretation of *primary* (/3), *secondary* (/6) or *uncertain* (/9) expressions depended on document-level cues. In contrast, differentiation grades (e.g., *grade 2*, *peu différencié*) were usually explicitly mentioned and required minimal contextual inference.

This reference to source context ensured that ICD-O codes were not only structurally correct, but also semantically accurate in the clinical narrative.

A custom set of Python scripts, based on and extending the open-source bratly package, supported the annotation, normalisation and translation workflow. These tools automated the extraction and reintegration of annotations, ICD-O dictionary matching, expression construction, and validation checks, as well as allowed integrative translation. Once the review process was complete, normalised codes were automatically propagated back into the .ann files, producing a consistent and fully enriched final dataset.

This combined methodology of manual annotation and programmatic tooling allowed for scalable, high-quality annotation across 1'301 documents and over 70'000 entities.

# Data Record

The dataset is available via the Zenodo repository at 10.5281/zenodo.17284817. It comprises 1'301 synthetic oncological clinical texts in French, derived from the CANTEMIST corpus and translated through the FRASIMED initiative. Each text file is provided in plain-text format (.txt) and is accompanied by annotation files (.ann) formatted according to the brat standoff annotation scheme. A .csv file of all annotations and ICD-O normalisation is also available for processing and review.

A total of 1'301 annotation files is included, containing over 70'000 manual annotations with ICD-O normalisation. These span 350 distinct morphology codes, 300 topography codes, all 4 differentiation codes, and more than 2'000 unique combinations expressed as expression_CIM annotations. Annotations are separated into entity annotations and corresponding normalisation notes referencing ICD-O terminology.

All files are stored in a flat directory structure. Each .txt file shares its filename prefix with its associated .ann file to denote correspondence (e.g., cc_onco859.txt, cc_onco859.ann). These filename prefixes correspond directly to the original CANTEMIST and FRASIMED corpus identifiers, enabling users to trace back and recover the source Spanish texts if needed for comparative or contextual analysis.

All scripts necessary to handle the data are available in the associated GitHub repository (https://github.com/SimedDataTeam/FRACCO).

The dataset is distributed under an open-access license intended for non-commercial use, with the requirement that this work is appropriately cited when reused.

# Technical Validation

To ensure the reliability and consistency of the dataset, we performed structured validation procedures across both the span annotation and ICD-O code normalisation phases. These two layers of validation reflect the dual nature of the annotation task: identifying relevant medical expressions in the text and assigning them appropriate standardised codes. In addition, we evaluated the dataset through NER model fine-tuning to confirm that the annotation scheme is coherent, the labels are learnable by modern architectures, and the corpus is of sufficient size to support supervised training. Together, these procedures confirm both the internal quality of the annotations and the usability of the dataset for NLP applications.

## Span Annotation Validation

The first stage involved validating text span annotations, highlighting expressions related to morphology, topography, and differentiation, across all 1,300 French clinical texts. Two biomedical expert annotators worked independently using a shared guideline that emerged from an initial calibration phase. During annotation, inconsistencies were noted particularly in non-pre-annotated terms, those not projected from the original CANTEMIST/FRASIMED dataset, which were harder to detect. To mitigate this, we implemented an automatic flagging mechanism that identified recurrent untagged terms, surfacing them in subsequent rounds. This significantly reduced cases where an annotation was completely missed by one annotator.

In the first round of full annotation, we computed both soft (partial overlap, *Figure 7*) and hard (exact match, *Figure 8*) inter-annotator agreement scores. Soft F1 scores ranged from 0.82 to 0.90, and hard F1 scores ranged from 0.70 to 0.90 across categories, with the exception of one outlier category (*differenciation*) which scored 0.48. Error analysis showed that most disagreements were due to span boundary differences: one annotator would often include more contextual information, while the other opted for a more concise span. The particularly low score for *differenciation* annotations was traced to a systematic discrepancy in span selection, one annotator consistently annotated expressions like "de grade 2," while the other selected only "grade 2," leading to a mismatch despite semantic equivalence.

For the other annotation categories, disagreements were largely due to different interpretations of what textual information was relevant to include, rather than divergent understandings of the clinical meaning. In addition, a significant number of "missing" annotations, where only one annotator marked an entity, were found to be non-pre-annotated terms that had been overlooked during initial review. These were flagged and subsequently reviewed by the annotator who had not originally captured them, improving overall coverage and consistency of the final corpus.

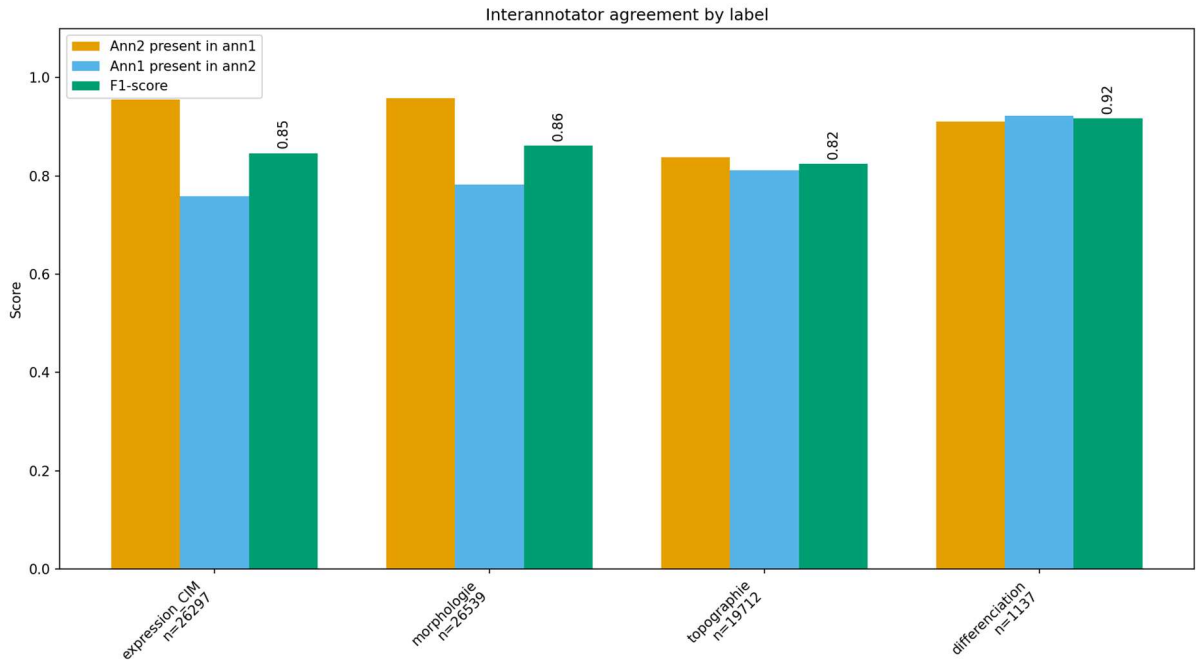

*Figure 7: inter-annotator agreement (partial agreement included)*

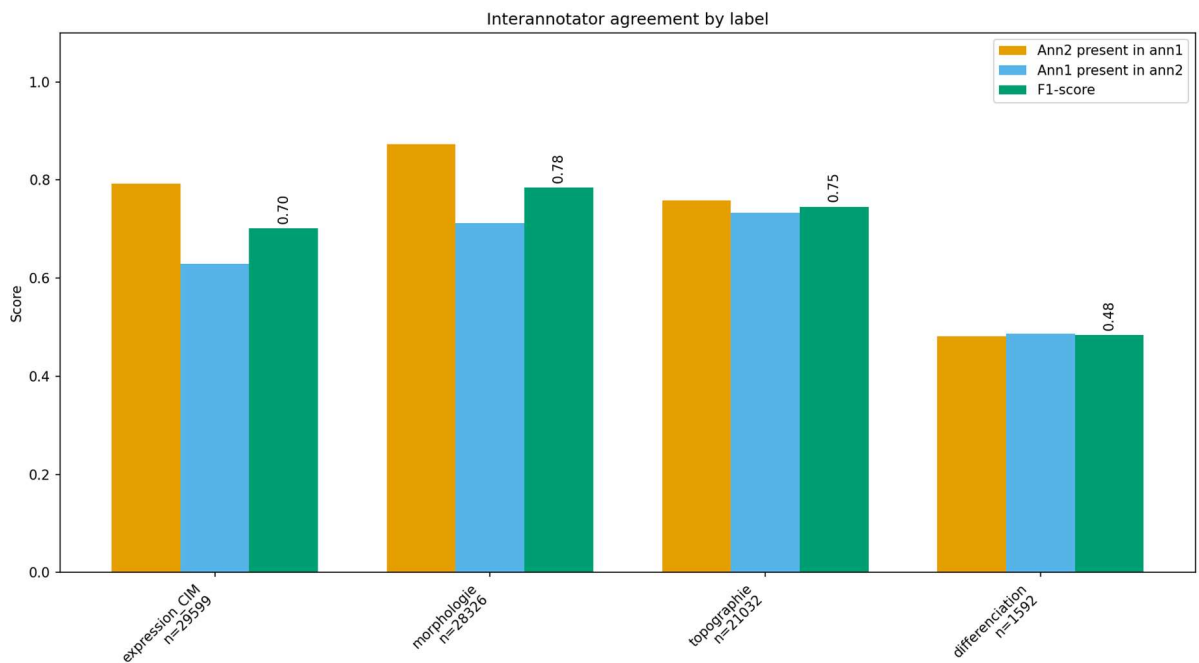

*Figure 8: inter-annotator agreement (perfect agreement only)*

For the final reconciliation phase, each annotator reviewed all partial-match cases. Annotations missing completely from one or the other annotator were reviewed and validated by a third annotator. Disagreements were resolved by consensus, and agreed annotations were included in the harmonised corpus; unresolved items default to the version provided by one annotator. This process yielded improved agreement, and a final, high-quality entity annotation set.

## ICD-O Normalisation Validation

For normalisation, annotated spans were exported into a structured CSV and matched via exact string matching against the ICD-O terminology dictionary. This method successfully normalised approximately 55'000 annotations (≈78%) automatically. The remaining ~15'000 complex expressions were manually reviewed. To evaluate both pathways, we conducted a stratified validation calculated to yield 95% confidence and 3% margin. From this, two validation corpora were extracted, an "automatic" and a "manual" corpus of respectively 1'060 and 1'000 expressions. Each corpus was then reviewed by a different expert, for codes to be validated.

In validating the automatic subset, the inter-annotator agreement (IAA) on the sampled 1,060 reached an 80.65% concordance. By contrast, validation of the manually assigned annotations revealed lower agreement, at 51.5%. This discrepancy highlights the increased difficulty in normalising expressions in the absence of direct dictionary matches.

Error analysis revealed several common sources of mismatch:

- Ontology limitations: ICD-O sometimes lacked precise codes for composite expressions, leading to multiple potential mappings, consistent with known challenges in ontology mapping (16–18)
- Semantic ambiguity: Different interpretations of clinical meaning led to variant code assignment decisions between annotators.
- Complex phrasing: Certain expressions combined elements in ways not directly supported by ICD-O, requiring consensus interpretation.

Annotators discussed all such edge cases, reached consensus coding, and propagated the agreed-upon mapping across the full dataset. A full list of the reviewed edge cases can be found in supplementary material.

Together, these validation steps demonstrate that our normalisation workflow, combining automated dictionary matching, expert review, and targeted error resolution, yields a high-quality annotated corpus suitable for downstream biomedical NLP applications.

## NER model fine-tuning on FRACCO

To demonstrate how this dataset can be leveraged for machine learning applications, we fine-tuned several pretrained named entity recognition (NER) models on the annotated texts. The goal was not to optimise model performance, but to provide first results to establish benchmark scores that enable comparison across future studies. Those results illustrate how different BERT-based models with varying pretraining corpora and scopes perform on the dataset.

The corpus, comprising 1,301 clinical case texts—a dataset size comparable to other French biomedical NER benchmarks (20)–was divided into a training set (80%) and a held-out test set (20%). The latter was kept fixed across all experiments and used only for final evaluation. Within the training portion, a validation set corresponding to 10% of the training data (8% of the full corpus) was resampled at each run, ensuring that training/validation splits varied while the test set remained constant. This design reduces dependence on any single validation partition while maintaining comparability. We fine-tuned five different BERT-based models, each trained with five fixed initialisation seeds.

Performance, averaged across seeds and evaluated with exact span and label matching, is summarised in Figures 9 and 10. Four of the five BERT-based models performed comparably, despite varying pretraining strategies. CamemBERT-base (pretrained on general-domain French texts drawn from the OSCAR Common Crawl corpus and French Wikipedia) and CamemBERT-bio (a continued-pretraining adaptation of CamemBERT-base on French biomedical corpora) both reached 89.4% weighted-F1. The multilingual general-domain models scored marginally lower, with multilingual BERT reaching 89.2% weighted-F1 and XLM-RoBERTa reaching 88.7%. By contrast, frALBERT underperformed at 85.3% weighted-F1, consistent with its smaller size, more limited pretraining (~4 GB of French Wikipedia), and efficiency-oriented design (21).

Differences among the four larger models were ≤1 F1 point, comparable to variation expected across random seeds. This contrasts with evaluations of French biomedical NER, where domain-adapted models such as CamemBERT-bio typically gain 2-3 F1 points over general-domain CamemBERT (20,22,23) and monolingual French encoders usually outperform multilingual BERT (24). In our case, no meaningful differences between model types were observed. This is likely explained by the fact that oncological terminology comprises frequent morphology and topography terms that are already well represented in general corpora such as OSCAR and Wikipedia, so additional biomedical adaptation contributes less than in tasks where vocabulary is rarer. The advantage of monolingual pretraining is also reduced in this setting, as the corpus provides a high density of short, repetitive entity mentions that can be effectively learned by both general-domain and multilingual models. Overall, the convergence of the larger models indicates that the dataset yields stable results across different pretrained variants.

At the entity level, *morphologie* achieved the highest F1 scores across models (90.5–91.1% for the larger models). *Topographie* and *différenciation* performed at comparable levels, with F1 scores ranging from 86.3-87.2% and 85.7-87.4%, respectively, despite the much smaller number of *différenciation* annotations. This is explained by the greater reproducibility of grade expressions, which are short and lexically homogeneous (e.g. "*de grade I*"). In contrast, *topographie* covers a wide lexical and semantic space. This is consistent with the ICD-O-3 structure, which defines only four differentiation codes compared to nearly 250 topography codes. Despite the strong imbalance across categories—*morphologie* and *topographie* together account for over 95% of annotations—F1 scores remained stable across all labels. This indicates that class imbalance did not hinder learnability, particularly for *différenciation*, which achieved high performance despite representing only ~2% of the data.

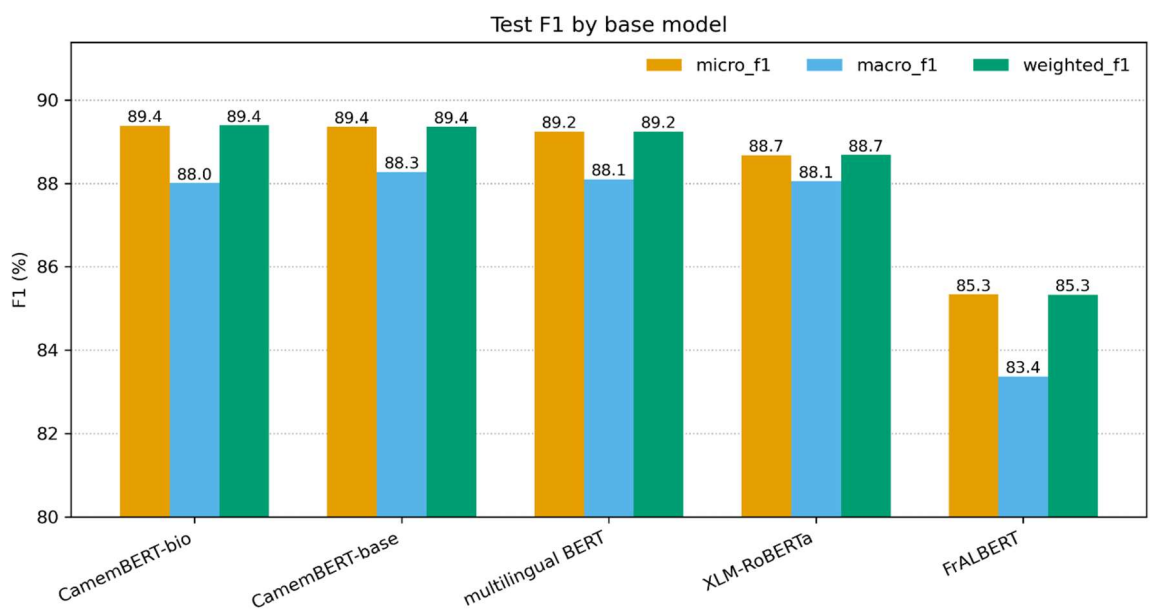

*Figure 9. Test set F1 scores (micro, macro and weighted) for five pretrained models fine-tuned on the corpus.*

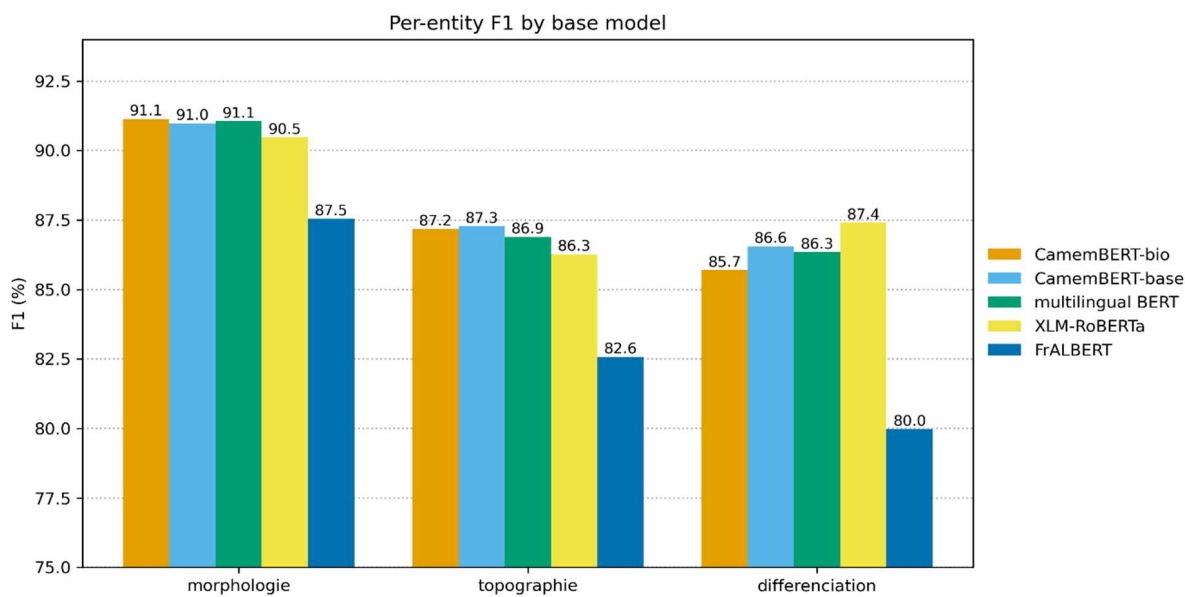

*Figure 10. Per-entity F1 scores (morphologie, topographie, différenciation) on the test set for five pretrained models fine-tuned on the corpus.*

# (Usage Notes)

The dataset is distributed as pairs of .txt and .ann files, following the BRAT annotation format. Each .txt file contains the raw French clinical text, and its corresponding .ann file contains the associated annotations. Annotations include both entity spans (for morphology, topography, and differentiation terms) and note-level annotations for ICD-O code normalisation, linked via annotation identifiers.

All files are stored in a flat directory structure, and each pair of .txt and .ann files share a common filename prefix (e.g., case001.txt and case001.ann). These filenames correspond to the original document identifiers from the CANTEMIST corpus, enabling researchers to cross-reference with the original Spanish-language data if needed.

To facilitate data reuse, a dedicated GitHub repository is provided alongside the dataset, containing Python scripts for:

- Parsing and loading the .ann files,
- Extracting and inspecting span and normalisation annotations,
- Converting the dataset into CSV format,
- Searching and filtering expressions based on ICD-O codes.

Researchers unfamiliar with BRAT or ICD-O coding may find these tools helpful for preprocessing or adapting the data to their own pipelines.

No additional preprocessing is required to access or analyse the data. However, users performing large-scale analyses or training machine learning models may wish to convert the annotations using the provided scripts or adapt them for their preferred text annotation schema.

# Code Availability

All code used to process, validate, and explore the dataset is available at a dedicated GitHub repository: https://github.com/SimedDataTeam/FRACCO. This includes tools for reading BRAT-formatted annotations, linking entities with their normalisations, and extracting composite expression_CIM annotations. A snapshot of the repository is also archived on Zenodo alongside the dataset 10.5281/zenodo.17284817 to ensure long-term accessibility and reproducibility.

# Acknowledgements

We would like to thank the team at the origin of the CANTEMIST dataset, as well as all the persons who worked on the FRASIMED corpus. Both teams allowed this project to build upon high-quality, openly available resources. We also thank the expert annotators for their careful work and the contributors who supported annotation tool development and data processing.

This work was funded by GTOP (*Geneva Translational Oncology Programme*) and the NOVA foundation (*Fondation pour l'Innovation sur le Cancer et la Biologie*, https://www.fondation-cancer.ch/), with help from the Leenards Foundation (https://leenaards.ch/) (J-P.G), the NCCR Evolving language (https://evolvinglanguage.ch/) (JZ, grant number #51NF40_180888), with a grant from the Private Foundation of the Geneva University Hospitals (https://www.fondationhug.org/), and the Placide Nicod Foundation (FA, grant number F02-13292). This project benefited from infrastructure provided by the HUG.

# Author information

## Authors and Affiliations

**Service des Sciences de l'Information Médicale (SIMED), Hôpitaux Universitaires de Genève, 4 rue Perret-Gentil, 1205 Genève, Suisse**

Johann Pignat, Milena Vucetic, Mina Bjelogrlic, Jamil Zaghir, Jean-Philippe Goldman, Christophe Gaudet-Blavignac, Christian Lovis

**Service d'Oncologie de Précisioin, Hôpitaux Universitaires de Genève, 4 rue Perret-Gentil, 1205 Genève, Suisse**

Johann Pignat, Fanny Amrein, Jonatan Bonjour, Michel Cuendet, Olivier Michielin

## Contributions

J.P., M.V. and M.B. collaborated on developing and establishing the methodology. J.Z., J-P.G., and C.G.-B. participated in technical support of the annotation infrastructure and were consulted as ontology specialists. J.P. and M.V. manually annotated, reviewed, and corrected the span of the entities. F.A. and J.B. participated in the elaboration and the review of ICD-O assignment, as well as the guidelines for the normalisation of edge cases J.P. annotated extracted expressions with ICD-O and SNOMED codes and corrected them after F.A. and J.B. reviews. Redaction of the article was conducted by J.P. with contributions from M.V., M.B., C.G-B. and M.C. C.L. and O.M. supervised the entire process and gave the final approval of the process. All authors reviewed and approved the final manuscript before submission

## Corresponding author

Correspondence to Johann Pignat

# Competing Interests

The authors declare that they have no known competing financial interests or personal relationships that could have influenced the work reported in this paper.